\documentclass[runningheads]{llncs}
\usepackage{graphicx}

\usepackage{tikz}
\usepackage{comment}
\usepackage{amsmath,amssymb} 
\usepackage{color}

\usepackage[accsupp]{axessibility}  

\usepackage{hyperref}

\usepackage[width=122mm,left=12mm,paperwidth=146mm,height=193mm,top=12mm,paperheight=217mm]{geometry}

\usepackage{booktabs}
\usepackage{multirow}

\begin{document}
\pagestyle{headings}
\mainmatter
\def\ECCVSubNumber{100}  

\title{Effectiveness of Function Matching in \\Driving Scene Recognition} 

%
\author{Shingo Yashima 
}
\authorrunning{Shingo Yashima}
%
\institute{Denso IT Laboratory, Inc.}
\maketitle

\begin{abstract}
Knowledge distillation is an effective approach for training compact recognizers required in autonomous driving. Recent studies on image classification have shown that matching student and teacher on a wide range of data points is critical for improving performance in distillation. This concept (called {\it function matching}) is suitable for driving scene recognition, where generally an almost infinite amount of unlabeled data are available. In this study, we experimentally investigate the impact of using such a large amount of unlabeled data for distillation on the performance of student models in structured prediction tasks for autonomous driving. Through extensive experiments, we demonstrate that the performance of the compact student model can be improved dramatically and even match the performance of the large-scale teacher by knowledge distillation with massive unlabeled data. 

\end{abstract}

\section{Introduction}
Recognition models in autonomous driving need to be lightweight and operate in real-time. On the other hand, there is a growing performance discrepancy between large-scale, state-of-the-art models and such lightweight models. This stems not only from the insufficient model capacity but also from the lack of the {\it inductive biases} in compact models: it has been well known that compact networks are harder to learn and generalize than large networks, even though they may have the capacity to represent the generalized solutions of large models. Therefore, establishing learning methods that exploit the full ability of compact models is crucial in this field.

Knowledge distillation \cite{hinton2015distilling,bucilua2006model} overcomes the lack of inductive biases by training a compact student model to emulate a large, generalized teacher model. Its superiority to standard supervised learning has been confirmed in numerous studies \cite{cho2019efficacy,Romero15fitnets}, and the state-of-the-art performance of a given architecture on typical vision tasks (e.g., ResNet-50 on ImageNet classification) is achieved by knowledge distillation \cite{beyer2022knowledge,ridnik2022solving}.
In the literature, these decent performances of knowledge distillation have been mainly attributed to the following two factors: 
\begin{enumerate}
    \item Soft labels or features maps produced by the large-scale teacher model contain rich semantic information (i.e., {\it dark knowledge} \cite{hinton2015distilling}). This leads the student to capture representations that generalize well \cite{muller2019does}.
    \item Student's output can be matched to the teacher's output on data point far from the original training data by using intensely augmented data or auxiliary unlabeled data. This allows the student to be more thoroughly signaled for generalization than supervised learning only with the labeled training data \cite{beyer2022knowledge}.
\end{enumerate}
Most of the existing studies mainly focused on the former factor and developed various distillation methods under a setting where same training data or augmentations as the supervised learning were applied for distillation \cite{furlanello2018born,lopez2015unifying}. However, recent studies \cite{beyer2022knowledge,ridnik2022solving,asano2021extrapolating} reveal that the latter factor is essential to achieving strong performance and simple output matching \cite{hinton2015distilling} on such massive data points exhibits state-of-the-art performance in image classification. This explanation on how knowledge distillation works is called {\it function matching} \cite{beyer2022knowledge} and its effectiveness is also discussed theoretically \cite{hsu2020generalization,cotter2021distilling}.

When considering learning compact recognizers for autonomous driving, the concept of function matching seems attractive: we can easily access an {\it almost infinite amount of} unlabeled driving scene data. From this viewpoint, matching the student's output to the teacher's output at such massive data points is expected to yield well-performed compact models. However, existing studies on knowledge distillation in structured scene recognition mainly focused on designing task-specific distillation losses \cite{li2017mimicking,liu2020structured} rather than quantitatively investigating the impact of increasing data points on which distillation is performed.

In this study, we examine the effectiveness of function matching in driving scene recognition by conducting extensive experiments on two fundamental tasks in this field: semantic segmentation and object detection. 
We found that the performance of the student distilled with a large unlabeled dataset not only exceeds the one by standard supervised learning but also matches the  performance of the teacher. This result indicates performance gap between teacher and student in supervised learning comes not from the difference in representational capacity but from that in inductive bias. In addition, we show that such a significant improvement is not observed when distilling only with original training data, indicating the number of distilling data points is crucial in function matching.

\subsubsection{Remark.}
Although many studies have investigated knowledge distillation methods suitable for structured prediction tasks \cite{li2017mimicking,liu2020structured} that utilize spatial relation of predictions or features, we focus on the plain output matching as we are interested in how well the student can match the teacher as a function. Somewhat surprisingly, the student can mimic the teacher sufficiently well with such a simple distillation when large unlabeled dataset is available.

In addition, we note that the use of an unlabeled dataset is slightly unusual, but is not novel, having been used effectively in previous studies \cite{hinton2015distilling,liu2020structured}. However, their usage is rather an additional option, and the size of their unlabeled dataset is not substantially larger than the original training dataset. We consider the setting where the number of unlabeled data is almost infinite and we can take (nearly) a new sample in each iteration, which is realistic in driving scene recognition and expected to be more effective.

\section{Related Work}
\subsubsection{Knowledge distillation.}
Knowledge distillation \cite{hinton2015distilling,bucilua2006model} transfers knowledge from a large-scale model to a compact model so as to improve the performance of compact networks. It has been applied to image classification by using the class probabilities produced from the large
model as targets for training the compact model \cite{hinton2015distilling,ba2014deep}
 or transferring the intermediate features \cite{Romero15fitnets,tian2019contrastive}.
 More recently, the concept of function matching was introduced, and it was shown that the original output matching \cite{hinton2015distilling} on various data points achieves state-of-the-art performance in ImageNet classification \cite{beyer2022knowledge,ridnik2022solving}.
 
 Besides image classification, several distillation methods have been proposed for structured prediction task, such as semantic segmentation \cite{Xie_undated-wb,liu2020structured} or object detection \cite{li2017mimicking,wang2019distilling,dai2021general}. They reported improved performance by exploiting spatial relationship of predictions or feature maps.
Our primary interest differs from these studies in that we focus on the impact of the dataset size used in distillation rather than the specific design of distillation loss. 

\subsubsection{Semi-supervised learning.}
The use of both labeled and unlabeled data can be seen as semi-supervised learning. 
Typical setting of semi-supervised learning is few-shot classification, where unlabeled data are utilized because labeled data are insufficient to learn a meaningful classifier \cite{miyato2018virtual,sohn2020fixmatch}.  
In contrast, we assume the size of labeled data is large enough to learn the strong teacher model and leverage unlabeled data for further exploiting the ability of compact student models (overcoming the lack of inductive biases \cite{cotter2021distilling}).

\section{Experiments}
We use BDD100K \cite{yu2020bdd100k} as a driving scene dataset. It consists of 100,000 driving videos, each 40 seconds long at 720p. Annotations are provided for each reference frame of 100,000 videos for object detection and 10,000 subset videos for semantic segmentation. We adapt these annotated data as an original training dataset for each task. For an auxiliary unlabeled dataset, we use frames cut from original video clips at 6 fps that result in 16,125,000 images, which is more than 1,000 times larger than the labeled dataset for segmentation. Note that train/val/test splits are common in original video clips and annotated datasets, so auxiliary unlabeled dataset does not contain val/test data of annotated datasets.

Throughout the experiments, we compare the performance of the following three training strategies:
\begin{enumerate}
    \item Directly train the compact model by supervised learning with the labeled dataset ({\bf Supervised}).
    \item First train the large model by supervised learning with the labeled dataset, then distill it to the compact model with the same labeled dataset ({\bf Distill}).
    \item First train the large model by supervised learning with the labeled dataset, then distill it to the compact model with the auxiliary unlabeled dataset ({\bf Distill-Aux}).
\end{enumerate}
Roughly speaking, we can see the effect of using soft targets instead of hard labels by comparing the performance of (1) and (2). Furthermore, we can see the impact of increasing matching data points by comparing the performance of (2) and (3) as the main interest in this study.

\subsection{Semantic Segmentation}
We use the segmentation architecture of PSPNet \cite{zhao2017pyramid} with a ResNet-101 and ResNet-18 \cite{he2016deep} as the teacher and the student network, respectively. For supervised learning, we adopt the standard pixel-wise cross entropy. 

\subsubsection{Distillation loss.}
For distillation in Distill and Distill-Aux, we consider a simple pixel-wise KL divergence loss:
\begin{equation}
\mathcal{L}_{distill} = \frac{1}{|\mathcal{I}|}\sum_{(x,y) \in \mathcal{I}} \sum_{c \in \mathcal{C}} \left[-p_{x, y, c}^T \log p_{x, y, c}^S + p_{x, y, c}^T \log p_{x, y, c}^T\right]
\end{equation}
where $p_{x, y, c}^T$ and $p_{x, y, c}^S$ represent the probability of class $c$ on the pixel coordinate $(x, y)$ produced by the teacher and the student, respectively. $\mathcal{I}$ is a set of pixel coordinates in prediction, and $\mathcal{C}$ is a set of classes. We do not use any additional loss term with respect to the original dataset's hard labels as in \cite{beyer2022knowledge}.

\subsubsection{Training setup.}
In all settings, we train segmentation networks from ImageNet pre-trained weights using stochastic gradient descent (SGD) with the momentum (0.9) and the weight decay (0.0001) with batch size 8. The learning rate is initialized to be 0.01 and is multiplied by $(1-\frac{iter}{max\_iter})^{0.9}$. For the number of training steps, we sweep over \{80K, 160K, 320K, 640K, 1280K\} as the optimal iteration number depends on training strategies. We apply standard augmentations, including random scaling (from 0.5 to 2.0) and random horizontal flipping, except for the Distill-Aux setting, where we have massive data and do not need to care about overfitting.
In all settings, patches of size 512 × 1024 are randomly cropped from (possibly augmented) images.

\begin{figure}
\centering
\includegraphics[width=0.55\linewidth]{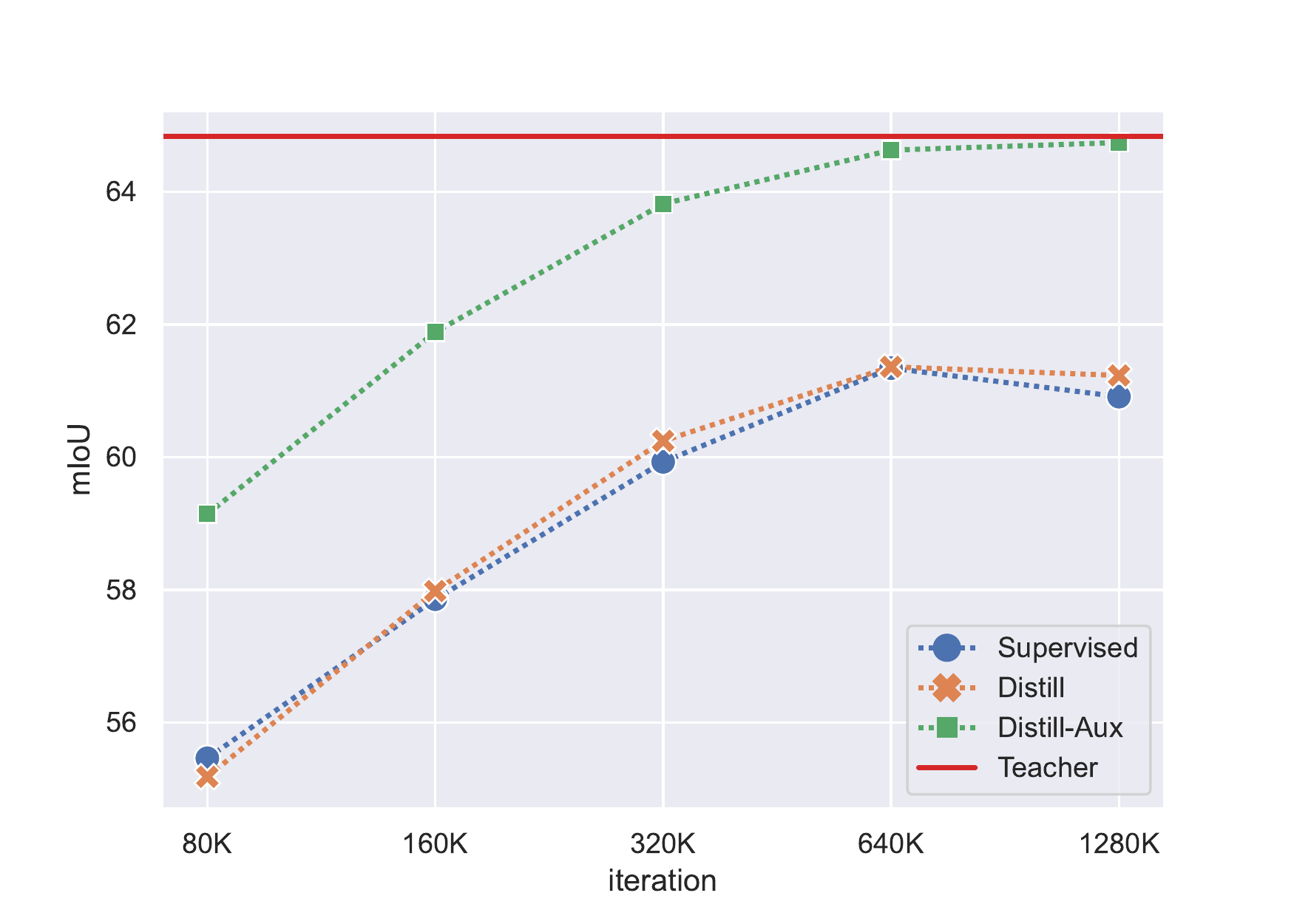}
\caption{mIoU results for semantic segmentation in all training steps. }
\label{fig:seg}
\end{figure}

\subsubsection{Results.}
The result in \autoref{fig:seg} shows that the compact model trained by Distill-Aux performs significantly better than the one trained by Supervised or Distill. This indicates that increasing the number of matching data points is the critical component  of the distillation, while the number of labeled data is the same for all training strategies.
\autoref{fig:seg_class} presents class-wise IoU of the best performing model among all training steps on each training strategy. While the improvement in major classes such as road or sky is relatively small, the significant performance gain is observed in more structured classes such as motorcycle or truck. This indicates that the semantic information in the teacher is transferred well to the student in Distill-Aux, thanks to the massive amount of matching data points.

\begin{figure}
\centering
\includegraphics[width=0.55\linewidth]{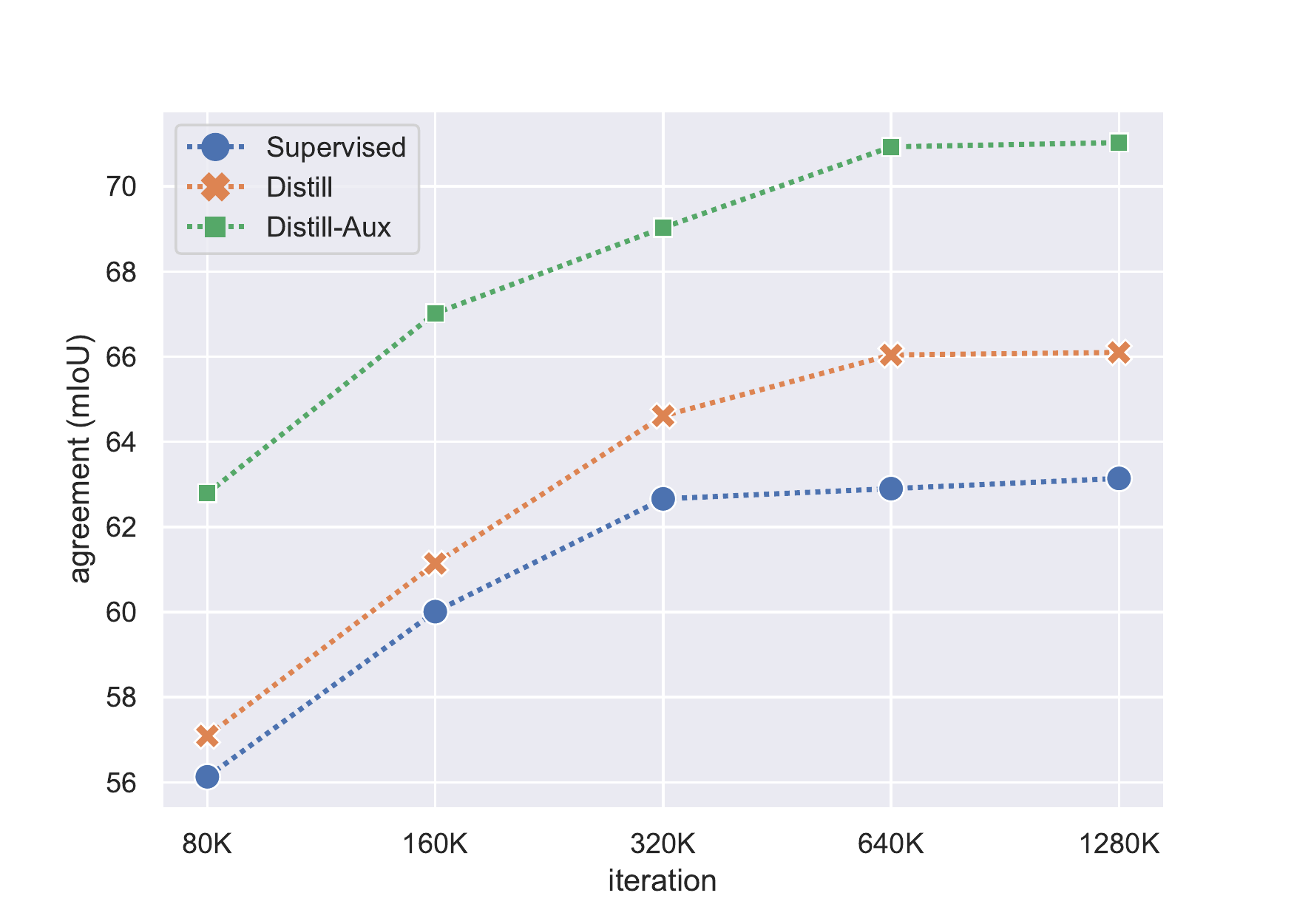}
\caption{Agreement between the teacher and the student predictions.}
\label{fig:seg_agr}
\end{figure}

In addition to the performance with respect to the ground truth, we show the prediction agreement between teacher and student in \autoref{fig:seg_agr}. Although it is natural that the agreement is low in Supervised, where we do not leverage the teacher's prediction, we can see Distill-Aux achieves much higher fidelity between the teacher and the student than Distill. This high fidelity to the teacher possibly leads to the large performance gain of Distill-Aux from Distill with respect to the ground truth in \autoref{fig:seg}.
 
\begin{table}[htbp]
\begin{center}
\caption{Class-wise IoUs for semantic segmentation. We pick the best-performing model among all training steps on each training strategy.}
\scalebox{0.88}{
\begin{tabular}{p{2.3cm}p{1.8cm}cccccccccc}
\toprule
Model & Method        & mIoU           & road           & swalk       & build.      & wall           & fence          & pole           & tlight  & tsign   & veg.     \\ 
\midrule
Teacher (R101)      & Supervised       &$64.83$           & $95.68$          & $70.68$          & $87.94$          & $28.97$          & $54.00$          & $58.30$          & $64.73$           & $63.83$          & $87.49$          \\ 
\midrule
\multirow{3}{*}{Student (R18)}      & Supervised       &$61.48$           & $94.84$          & $66.76$          & $86.42$          & $\bf{30.84}$          & $48.52$          & $52.04$          & $59.99$           & $58.67$          & $86.43$          \\ 
     &Distill & $61.70$           & $94.96$ & $66.60$ & $86.44$ &$25.75$ & $48.89$ &$ 51.66$          &$58.82$ & $58.89$ & $86.55$ \\ 
     &Distill-Aux    & $\bf{64.74} $          & $\bf{95.59}$          & $\bf{71.11}$          & $\bf{87.92}$           & $28.80$          & $\bf{54.00}$          & $\bf{54.66}$ & $\bf{62.59}$          & $\bf{61.43}$          & $\bf{87.82}$          \\ 
\bottomrule
\toprule
Model & Method            & terrain        & sky            & person         & rider          & car            & truck          & bus            & train          & mbike    & bike        \\ 
\midrule
Teacher (R101)     &Supervised       &$51.51$           & $95.59$          & $69.00$          & $36.33$          & $91.55$          & $58.12$          & $83.60$          & $0.00$           & $70.80$          & $63.57$          \\ 
\midrule
\multirow{3}{*}{Student (R18)}      &Supervised        & $47.63$          & $94.91$          & $65.10$          & $35.54$          & $90.68$          & $54.17$          & $80.54$          & $0.00$          & $56.38$          & $58.63$          \\ 
     &Distill & $48.02$ & $94.92$ & $64.56$  & $41.61$ & $90.82$          & $56.45$          & $81.16$          & $0.00$          & $57.40$          & $58.82$          \\ 
     &Distill-Aux      & $\bf{51.96}$          & $\bf{95.71}$          & $\bf{68.76}$           & $\bf{44.95}$          &$ \bf{92.04}$  & $\bf{62.75}$ & $\bf{83.34}$ & $0.00$ & $\bf{63.23}$ & $\bf{63.32}$ \\ 
\bottomrule
\end{tabular}
}
\label{fig:seg_class}
\end{center}
\vskip -0.1in
\end{table}

\subsection{Object Detection}
As single stage detectors are easy to implement output matching, we use the object detection architecture of FCOS \cite{tian2020fcos} with a ResNet-101 and ResNet-18 \cite{he2016deep} as the teacher and the student network, respectively. The channel in the detector towers is set to 128 for the student and 256 for the teacher. For supervised learning, we adapt the original loss of FCOS.

\subsubsection{Distillation loss.}
We use the following loss function for output matching, which is a natural extension of the original loss \cite{tian2020fcos} for soft targets:
\begin{align}
    &\mathcal{L}_{cls} = \frac{1}{N_{pos}}\sum_{(x,y) \in \mathcal{I}} \sum_{c \in \mathcal{C}} \left[-p_{x, y, c}^T \log p_{x, y, c}^S - (1-p_{x, y, c}^T) \log (1-p_{x, y, c}^S)\right], \\
    &\mathcal{L}_{reg} = \sum_{(x,y) \in \mathcal{I}} \frac{\sum_{c \in \mathcal{C}} p_{x, y, c}^T}{N_{pos}}\mathrm{IoU}(t^T_{x, y}, t^S_{x, y}),\\
    &\mathcal{L}_{center} = \sum_{(x,y) \in \mathcal{I}} \left[-q_{x, y}^T \log q_{x, y}^S - (1-q_{x, y}^T) \log (1-q_{x, y}^S)\right],\\
    &\mathcal{L}_{distill} = \mathcal{L}_{cls} + \mathcal{L}_{reg} + \mathcal{L}_{center}
\end{align}
where $p_{x, y, c}$, $t_{x, y}$, and $q_{x, y}$ are the outputs for class probability, regression coordinate, and centerness, respectively. $\mathcal{I}$ is a set of pixel coordinates in the prediction of all feature levels, and $\mathcal{C}$ is a set of classes. $N_{pos} = \sum_{(x,y) \in \mathcal{I}} \sum_{c \in \mathcal{C}} p_{x, y, c}^T$, and $\mathrm{IoU}$ is the GIoU loss as in \cite{yu2016unitbox}.
When replacing the teacher's output with hard label, we recover the original FCOS loss.

\subsubsection{Training setup.}
Similar to semantic segmentation, we train detection networks from ImageNet pre-trained weights using stochastic gradient descent (SGD) with the momentum (0.9) and the weight decay (0.0001) with batch size 32. The learning rate is initialized to be 0.01 and is multiplied by $(1-\frac{iter}{max\_iter})^{0.9}$. For the number of training steps, we sweep over \{80K, 160K, 320K, 640K\}. We apply standard augmentations, including random scaling (from 0.5 to 2.0) and random horizontal flipping, except for the Distill-Aux setting as in semantic segmentation.

\subsubsection{Results.}
\begin{figure}
\centering
\includegraphics[width=0.55\linewidth]{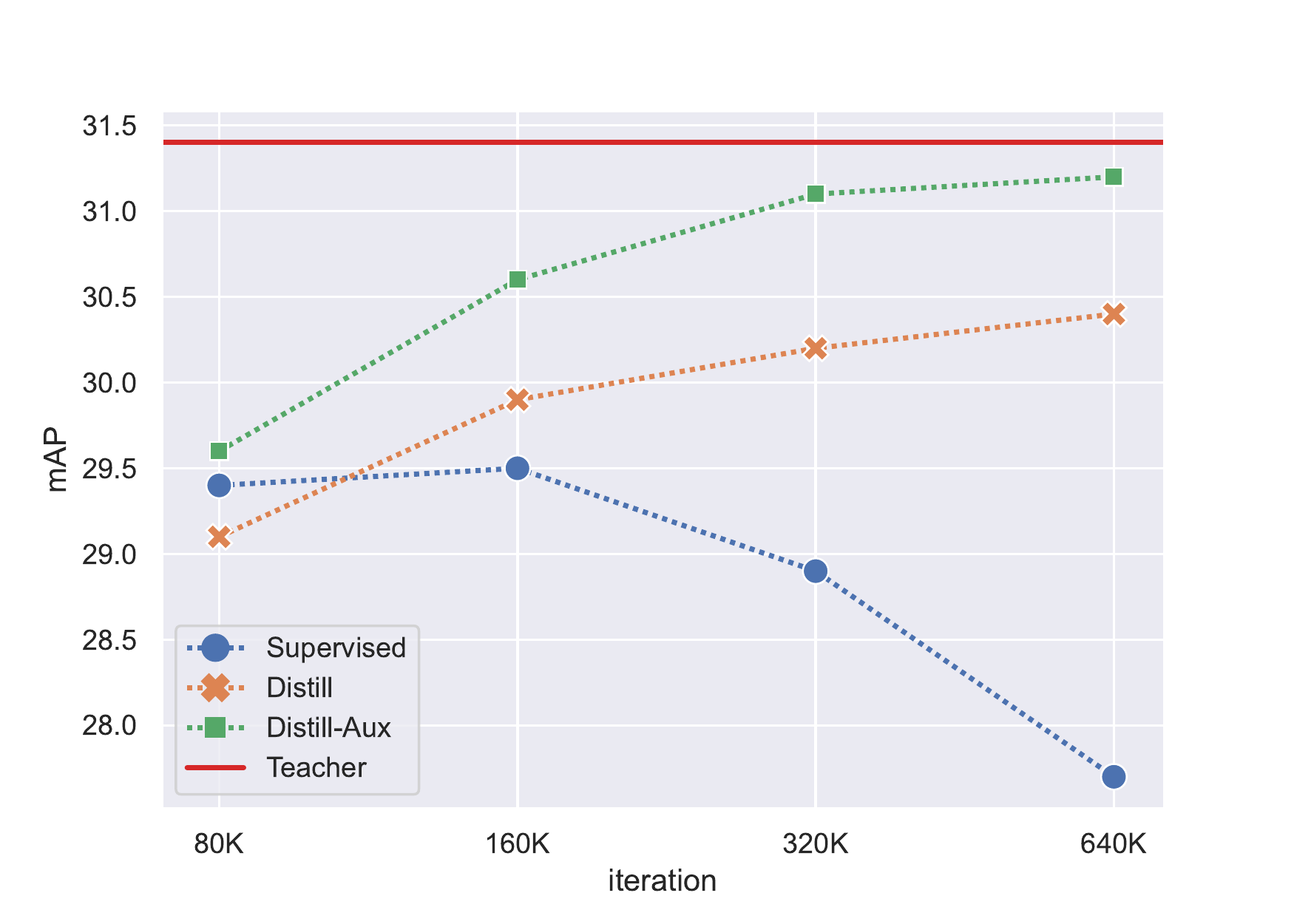}
\caption{mAP results for object detection in all training steps. }
\label{fig:det}
\end{figure}
Similar to the results of semantic segmentation, \autoref{fig:det} shows that the compact model trained by Distill-Aux performs the best and almost reaches the teacher's performance. In contrast, the model trained by Supervised falls into overfitting in large training steps.
From the detailed scores shown in \autoref{fig:det_acc}, we can see that Distill-Aux improves performance in all metrics, especially for large objects (APl). This might indicate that the performance gain comes from the improvements in classification rather than the regression quality.

\begin{table}[htbp]
\begin{center}
\caption{Detailed results for object detection. We pick the best-performing model among all training steps on each training strategy.}
\scalebox{0.90}{
\begin{tabular}{p{2.3cm}p{2cm}cccccc}
\toprule
Model & Method        & \quad mAP \quad           & \quad AP50 \quad           & \quad AP75  \quad      & \quad APs  \quad    & \quad  APm    \quad        & \quad APl   \quad         \\ 
\midrule
Teacher (R101) & Supervised       &\quad $31.4$   \quad         & \quad $56.3$  \quad         & \quad $29.7$   \quad        & \quad $14.6$   \quad        & \quad $37.9$   \quad        & \quad $52.1$    \quad          \\ 
\midrule
 \multirow{3}{*}{Student (R18)} &Supervised       &\quad $29.5$  \quad          &\quad  $53.9$      \quad     & \quad  $27.4$   \quad        & \quad $13.5$  \quad         & \quad $34.4$   \quad        & \quad  $48.4$    \quad         \\ 
 &Distill & \quad $30.4$    \quad        & \quad $55.1$ \quad  & \quad $28.3$ \quad & \quad $\bf{13.7}$ \quad & \quad $35.2$ \quad  & \quad $51.0$ \quad  \\ 
 &Distill-Aux      & \quad $\bf{31.2} $ \quad          & \quad $\bf{56.3}$      \quad     & \quad $\bf{29.0}$ \quad           & \quad $\bf{13.7}$   \quad         & \quad $\bf{36.3}$  \quad         & \quad $\bf{52.6}$   \quad          \\ 
\bottomrule
\end{tabular}
}
\label{fig:det_acc}
\end{center}
\vskip -0.1in
\end{table}

\section{Conclusion}
In this study, we experimentally investigated the influence of the amount of unlabeled data used in distillation for structured predictions. We confirmed that the use of large unlabeled data significantly improves compact models' performance. We believe these results are promising for autonomous driving since collecting such a massive amount of unlabeled data is easy in this domain.
Although the type of the backbone architecture used for teachers and students was identical (ResNet) in this study, it is an important issue to investigate the effectiveness of function matching in non-identical settings \cite{somepalli2022can} (e.g., transformer teacher and ResNet student) because compatibility with hardware heavily depends on these architectures. This might give more principled strategies for learning real-time models for driving scene recognition.

%
%
\bibliographystyle{splncs04}
\bibliography{egbib}
\end{document}